%% file: main.tex
\documentclass[10pt]{article}

\usepackage{graphicx}
\usepackage{amssymb,amsfonts,amsmath}
\usepackage{array}

\usepackage{cite}
\usepackage{color} 

\usepackage[labelfont=bf,labelsep=period,justification=raggedright]{caption}

\makeatletter
\renewcommand{\@biblabel}[1]{\quad#1.}
\makeatother

\date{}

\usepackage{url}

\newcommand{\change}[1]{#1}

\begin{document}

\begin{flushleft}
{\Large
\textbf{Diffusion of Lexical Change in Social Media}
}
\\
Jacob Eisenstein$^{1,\ast}$, 
Brendan O'Connor$^{2}$, 
Noah A. Smith$^{3}$,
Eric P. Xing$^{3}$
\\
\bf{1} School of Interactive Computing, Georgia Institute of Technology, Atlanta GA, USA
\\
\bf{2} School of Computer Science, University of Massachusetts, Amherst MA, USA
\\
\bf{3} School of Computer Science, Carnegie Mellon University, Pittsburgh PA, USA
\\
$\ast$ E-mail: jacobe@gatech.edu
\end{flushleft}

\section*{Abstract}
Computer-mediated communication is driving fundamental changes in the nature of written language.  We investigate these changes by statistical analysis of a dataset comprising 107 million Twitter messages (authored by 2.7 million unique user accounts).  Using a latent vector autoregressive model to aggregate across thousands of words, we identify high-level patterns in diffusion of linguistic change over the United States. Our model is robust to unpredictable changes in Twitter's sampling rate, and provides a probabilistic characterization of the relationship of macro-scale linguistic influence to a set of demographic and geographic predictors. The results of this analysis offer support for prior arguments that focus on geographical proximity and population size. However, demographic similarity -- especially with regard to race -- plays an even more central role, as cities with similar racial demographics are far more likely to share linguistic influence. Rather than moving towards a single unified ``netspeak'' dialect, language evolution in computer-mediated communication reproduces existing fault lines in spoken American English.

\section*{Introduction}
An increasing proportion of informal communication is conducted in
written form, mediated by technology such as smartphones and social
media platforms. Written language has been forced to adapt to meet the demands of synchronous conversation, resulting in a creative burst of new forms, such as emoticons, abbreviations, phonetic spellings, and other neologisms~\cite{androutsopoulos2000non,anis2007neography,herring2012grammar}. Such changes have often been considered as a single, uniform dialect --- both by researchers~\cite{crystal2006language,squires2010enregistering} and throughout the popular press~\cite{thurlow2006from,squires2010enregistering}. But despite the fact that social media facilitates instant communication between distant corners of the earth, the adoption of new written forms is often sharply delineated by geography and demographics~\cite{eisenstein2010latent,eisenstein2011discovering,schwartz2013personality}.
For example, \change{in our corpus of social media text from 2009 to 2012}, the abbreviation \textit{ikr} (\textit{I know, right?})
occurs six times more frequently in the Detroit area than in the
United States overall; the emoticon
{\textasciicircum}-{\textasciicircum} occurs four times more
frequently in Southern California; the phonetic spelling
\textit{suttin} (\textit{something}) occurs five times more
frequently in New York City.

These differences raise questions about how language change spreads in
online communication.  What groups are influential, and which communities evolve together?  Is written language moving toward global standardization or increased fragmentation? As language is a crucial constituent of personal and group identity, examination of the
competing social factors that drive language change can shed new light
on the hidden structures that shape society. This paper offers a new technique for inducing networks of linguistic influence and co-evolution from raw word counts. We then seek explanations for this network in a set of demographic and geographic predictors, using a logistic regression in which these predictors are used to explain the induced transmission pathways.

A wave of recent research has shown how social media datasets can enable large-scale analysis of patterns of communication~\cite{lotan2011arab,wu2011says}, sentiment~\cite{dodds2011temporal,thelwall2009homophily,mitchell2013geography}, and influence~\cite{lazer2009computational,aral2012identifying,bond2012million,gomez2012inferring,bakshy2012role}. Such work has generally focused on tracking the spread of discrete
behaviors, such as using a piece of software~\cite{aral2012identifying}, reposting duplicate or near-duplicate content~\cite{leskovec2009meme,cha2010measuring,lotan2011arab}, voting in political elections~\cite{bond2012million}, or posting a hyperlink to online content~\cite{gomez2012inferring,bakshy2012role}. Tracking linguistic changes poses a significant additional challenge, as we are concerned not with the first appearance of a word, but with the bursts and lulls in its popularity over time~\cite{altmann2009beyond}.  In addition, the well known ``long-tail'' nature of both word counts and city sizes~\cite{zipf1949human} ensures that most counts for words and locations will be sparse, rendering simple frequency-based methods inadequate. 

Language change has long been an active area of research, and a variety of theoretical models have been proposed. In the 
\emph{wave} model, linguistic innovations spread through interactions over the course of an individual's life, so the movement of linguistic innovation from one region to another depends on the density of interactions~\cite{bailey1973variation}.  In the simplest version of this model, the probability of contact between two individuals depends on their distance, so linguistic innovations should diffuse continuously through space.  \change{The \emph{gravity} model combines population and geographical distance: starting from the premise that the likelihood of contact between individuals from two cities depends on the size of the cities as well as their distance, this model predicts that linguistic innovations will travel between large cities first~\cite{trudgill1974linguistic}. The closely-related \emph{cascade} model focuses on differences in population, arguing that linguistic changes will proceed from the largest cities to the next largest, passing over sparsely populated intermediate geographical areas~\cite{labov2003pursuing}. Quantitative validation of these models has focused on edit-distance metrics of pronunciation differences amongst European dialects, with mixed findings on the relative importance of geography and population~\cite{nerbonne2007geographic,heeringa2007geografie,nerbonne2010measuring}.}

Cultural factors also play an important role in both the diffusion of, and resistance to, language change. Many words and phrases have entered the standard English lexicon from minority dialects~\cite{lee1999out}; conversely, there is evidence that minority groups in the United States resist regional sound changes associated with European American speakers~\cite{gordon2000phonological}, and that racial differences in speech persist even in conditions of very frequent social contact~\cite{rickford1985ethnicity}. At present there are few quantitative sociolinguistic accounts of how geography and demographics interact~\change{\cite{wieling2011quantitative}}; nor are their competing roles explained in the menagerie of theoretical models of language change, such as evolutionary biology~\cite{zhang2013principles,baxter2006utterance}, dynamical systems~\cite{niyogi1997dynamical}, Nash equilibria~\cite{trapa2000nash}, Bayesian learners~\cite{reali2010words}, and agent-based simulations~\cite{fagyal2010centers}. In general, such research is concerned with demonstrating that a proposed theoretical framework can account for observed phenomena like geographical distribution of
linguistic features and their rate of adoption over time. In contrast,
this paper \change{takes a data-driven approach}, fitting a model to a large corpus of text data from individual language users, and analyzing the social meaning of the resulting parameters. 

Research on reconstructing language phylogenies from cognate tables is also related~\cite{gray2003language,gray2009language,bouckaert2012mapping,dunn2011evolved}, but rather than a phylogenetic process in which languages separate and then develop in relative independence, we have closely-related varieties of a single language, which are in constant interaction. Other researchers have linked databases of typological linguistic features (such as morphological complexity) with geographical and social properties of the languages' speech communities~\cite{lupyan2010language}. Again, our interest is in more subtle differences within the same language, rather than differences across the entire set of world languages. The typological atlases and cognate tables that are the basis such work are inapplicable to our problem, requiring us to take a corpus-based approach~\cite{szmrecsanyi2011corpus}, estimating an influence network directly from raw text.

The overall aim of this work is to build a computational model capable of identifying the demographic and geographic factors that drive the spread of newly popular words in online text. To this end, we construct a statistical procedure for recovering networks of linguistic diffusion 
from raw word counts, even as the underlying social media sampling rate changes unaccountably. We present a procedure for Bayesian inference in this model, capturing uncertainty about the induced diffusion network. We then consider a range of demographic and geographic factors that might explain the networks induced from this model, using a post hoc logistic regression analysis. This lends support to prior work on the importance of population and geography, but reveals a strong role for racial homophily at the level of city-to-city linguistic influence.

\section*{Materials and methods}
We conducted a statistical analysis of a corpus of public data from the microblog site Twitter, from 2009--2012. The corpus includes 107 million messages, mainly in English, from more than 2.7 million unique user accounts. Each message contains GPS coordinates to locations in the continental United States. The data was temporally aggregated into 165 week-long bins. After taking measures to remove marketing-oriented accounts, each user account was associated with one of the 200 largest Metropolitan Statistical Areas (MSA) in the United States, based on their geographical coordinates. The 2010 United Census provides detailed demographics for MSAs. By linking this census data to changes in word frequencies, we can obtain an aggregate picture of the role of demographics in the diffusion of linguistic change in social media.

Empirical research suggests that Twitter's user base is younger, more urban, and more heavily composed of ethnic minorities, in comparison with the overall United States population~\cite{mislove2011understanding,duggan2013social}. Our analysis does \emph{not} assume that Twitter users are a representative demographic sample of their geographic areas. Rather, we assume that on a macro scale, the diffusion of words between metropolitan areas depends on the overall demographic properties of those areas, and not on the demographic properties specific to the Twitter users that those areas contain. Alternatively, the use of population-level census statistics can be justified on the assumption that the demographic skew introduced by Twitter --- for example, towards younger individuals --- is approximately homogeneous across cities. Table~\ref{tab:avg-demo} shows the average demographics for the 200 MSAs considered in our study.

Linguistically, our analysis begins with the 100,000 most frequent terms overall. We narrow this list to 4,854 terms whose frequency changed significantly over time; the excluded terms have little dynamic range; they would therefore not substantially effect on the model parameters, but would increase the computational cost if included. We then manually refine this list to 2,603 English words, by excluding names, hashtags, and foreign language terms. A complete list of terms is given in Appendix S1, examples of each term are given in Appendix S2, and more detailed procedures for data acquisition are given in Appendix S3. Manual annotations of each term are given in Table S1, and the software for our data preprocessing pipeline is given in Software S1.

Figure~\ref{f:ex-bruh} shows the geographical distribution of six words over time. The first row shows the word \textit{ion}, which is a shortened form of \textit{I don't}, as in \textit{ion even care}. Systematically coding a random sample of 300 occurrences of the string
\textit{ion} in our dataset revealed two cases of the
traditional chemistry sense of \textit{ion}, and 294 cases that clearly matched \textit{I don't}. This word displays increasing popularity over time, but remains strongly associated with the Southeast. In contrast, the second row shows the emoticon \textit{-\_\_-} (indicating annoyance), which spreads from its initial bases in coastal cities to nationwide popularity.  The third row shows the abbreviation \textit{ctfu}, which stands for \textit{cracking the fuck up} (i.e.,~laughter). At the beginning of the sample it is active mainly in the Cleveland area; by the end, it is widely used in Pennsylvania and the mid-Atlantic, but remains rare in the large cities to the west of Cleveland, such as Detroit and Chicago. What explains the non-uniform spread of this term's popularity?

While individual examples are intriguing, we seek an aggregated account of the spatiotemporal dynamics across many words, which we can correlate against geographic and demographic properties of metropolitan areas. Due to the complexity of drawing inferences about influence and demographics from raw word counts, we perform this process in stages. A block diagram of the procedure is shown in Figure~\ref{fig:block-diagram}. First, we model word frequencies as a dynamical system, using Bayesian inference over the latent spatiotemporal activation of each word. We use sequential Monte Carlo~\cite{Godsill2004Monte} to approximate the distribution over spatiotemporal activations with a set of samples. Within each sample, we induce a model of the linguistic dynamics between metropolitan areas, which we then discretize into a set of pathways. Finally, we perform logistic regression to identify the geographic and demographic factors that correlate with the induced linguistic pathways. By aggregating across samples, we can estimate the confidence intervals of the resulting logistic regression parameters. 

\subsection*{Modeling spatiotemporal lexical dynamics in social media data}
This section describes our approach for modeling lexical dynamics in our data. We represent our data as counts $c_{w,r,t}$, which is the number of individuals who used the word $w$ at least once in MSA $r$ at time $t$ (i.e., one week). (Mathematical notation is summarized in Table~\ref{tab:notation}. We do not consider the total number of times a word is used, since there are many cases of a single individual using a single word hundreds or thousands of times.) To capture the dynamics of these counts, we employ a latent vector autoregressive model, based on the binomial distribution with a logistic link function. The use of latent variable modeling is motivated by properties of the data that are problematic for simpler autoregressive models that operate directly on word counts and frequencies (without a latent variable). We begin by briefly summarizing these problems; we then present our model, describe the details of inference and estimation, and offer some examples of the inferences that our model supports.

\subsubsection*{Challenges for direct autoregressive models}
The simplest modeling approach would be an autoregressive model that operates directly on the word counts or frequencies~\cite{wei1994time}. A major challenge for such models is
that Twitter offers only a sample of all public messages, and the sampling rate can change in unclear ways~\cite{morstatter2013sample}. 
For example, for much of the timespan of our data, Twitter's documentation implies that the sampling rate is approximately 10\%; but in 2010 and earlier, the sampling rate appears to be 15\% or 5\%. (This estimate is based on inspection of message IDs modulo 100, which appears to be how sampling was implemented at that time.) After 2010, the volume growth in our data is relatively smooth, implying that the sampling is fair (unlike findings of \cite{morstatter2013sample}, which focus on a more problematic case involving query filters, which we do not use).

Raw counts are not appropriate for analysis, because the MSAs have wildly divergent numbers of users and messages. New York City has four times as many active users as the 10th largest metropolitan area (San Francisco-Oakland, CA), twenty times as many as the 50th largest (Oklahoma City, OK), and 200 times as many as the 200th largest (Yakima, WA); these ratios are substantially larger when we count messages instead of active users. This necessitates normalizing the counts to frequencies $p_{w,r,t} = c_{w,r,t} / s_{r,t}$, where $s_{r,t}$ is the number of individuals who have written at least one message in region $r$ at time $t$. The resulting frequency $p_{w,r,t}$ is the empirical probability that a random user in $(r,t)$ used the word $w$. Word frequencies treat large and small cities more equally, but suffer from several problems:
\begin{itemize}
  \item The frequency $p_{w,r,t}$ is \emph{not} invariant to a change in the sampling rate: if, say, half the messages are removed, the probability of seeing a user use any particular word goes down, because $s_{r,t}$ will decrease more slowly than $c_{w,r,t}$ for any $w$. The changes to the global sampling rate in our data drastically impact $p_{w,r,t}$.
\item Users in different cities can be more or less actively engaged with Twitter: for example, the average New Yorker contributed 55 messages to our dataset, while the average user within the San Francisco-Oakland metropolitan area contributed 21 messages. Most cities fall somewhere in between these extremes, but again, this ``verbosity'' may change over time. 
\item Word popularities can be driven by short-lived global phenomena, such as holidays or events in popular culture (e.g.,~TV shows, movie releases), which are not interesting from the perspective of persistent changes to the lexicon. We manually removed terms that directly refer to such events (as described in the Appendix S3), but there may be unpredictable second-order phenomena, such as an emphasis on words related to outdoor cooking and beach trips during the summer, and complaints about boredom during the school year.
\item Due to the long-tail nature of both word counts and city populations~\cite{clauset2009power}, many word counts in many cities are zero at any given point in time. This floor effect means that least squares models, such as Pearson correlations or the Kalman smoother, are poorly suited for this data, in either the $c_{w,r,t}$ or $p_{w,r,t}$ representations.
\end{itemize}

\subsubsection*{Latent vector autoregressive model}
To address these issues, we build a latent variable model that controls for these confounding effects, yielding a better view of the underlying frequency dynamics for each word. Instead of working with raw frequencies $p_{w,r,t}$, we perform inference over latent variables $\eta_{w,r,t}$, which represent the underlying \emph{activation} of word $w$ in MSA $r$ at time $t$. \change{We can convert between these two representations using the logistic transformation, $p_{w,r,t} = \textrm{Logistic}(\eta_{w,r,t})$, where
$\textrm{Logistic}(\eta) = 1/(1+e^{-\eta})$. We will estimate each $\eta_{w,r,t}$ by maximizing the likelihood of the observed count data $c_{w,r,t}$, which we treat as a random draw from a binomial distribution, with the number of trials equal to $s_{r,t}$, and the frequency parameter equal to $\text{Logistic}(\eta_{w,r,t})$.}

An $\eta$-only model, therefore, would be
\begin{equation}
  \label{e:basic}
  c_{w,r,t} \sim \textrm{Binomial}(s_{r,t}, \textrm{Logistic}(\eta_{w,r,t}))
\end{equation}
\noindent
This is a very simple generalized linear model with a logit link function~\cite{Gelman2006ARM}, in which the maximum likelihood estimate of $\eta$ would simply be a log-odds reparameterization of the probability of a user using the word, $\hat{\eta}_{w,r,t}=\log(p_{w,r,t}/(1-p_{w,r,t}))$.
By itself, this model corresponds to directly using $p_{w,r,t}$, and has all the same problems as noted in the previous section; in addition, the estimate $\hat{\eta}_{w,r,t}$ goes to negative infinity when $c_{w,r,t} = 0$.

The advantage of the logistic binomial parameterization is that it allows an additive combination of effects to control for confounds.
To this end, we include two additional parameters $\nu_{w,t}$ and $\mu_{r,t}$:
\begin{equation}
\label{e:observation}
c_{w,r,t} \sim \textrm{Binomial}(s_{r,t}, \textrm{Logistic}(\eta_{w,r,t} + \nu_{w,t} + \mu_{r,t})).
\end{equation}
The parameter $\nu_{w,t}$ represents the overall activation of the word $w$ at time $t$, thus accounting for non-geographical changes, such as when a word becomes more popular everywhere at once. The parameter $\mu_{r,t}$ represents the ``verbosity'' of MSA $r$ at time $t$, which varies for the reasons mentioned above. These parameters control for global effects due to $t$, such as changes to the API sampling rate. (Because $\mu_{r,t}$ and $\nu_{w,t}$ both interact with $t$, it is unnecessary to introduce a main effect for $t$.) 
\change{In this model, the $\eta$ variables still represent differences in log-odds, but after controlling for ``base rate'' effects; they can be seen an adjustment to the base rate, and can be estimated with greater stability.}

We can now measure lexical dynamics in terms of the latent variable $\eta$ rather than the raw counts $c$. We take the simplest possible approach, modeling $\eta$ as a first-order linear dynamical system with Gaussian noise~\cite{gelb1974applied},
\vspace{-.1in}
\begin{equation}
\eta_{w,r,t} \sim N\left(\sum_{r'} a_{r',r} \eta_{w,r',t-1}, \sigma^2_{w,r}\right).
\label{e:dynamics}
\end{equation}

The dynamics matrix $A = \{a_{r_1,r_2}\}$ is shared over both words and time; we also assume homogeneity of variance within each metropolitan area (per word), using the variance parameter $\sigma^2_{w,r}$. These simplifying assumptions are taken to facilitate statistical inference, by keeping the number of parameters at a reasonable size. If it is possible to detect clear patterns of linguistic diffusion under this linear homoscedastic model, then more flexible models should show even stronger effects, if they can be estimated successfully; we leave this for future work. It is important to observe that this model \emph{does} differentiate directionality: in general, $a_{r_1,r_2} \neq a_{r_2,r_1}$. The coefficient $a_{r_1,r_2}$ reflects the extent to which $\eta_{r_1,t}$ predicts $\eta_{r_2,t+1}$, and vice versa for $a_{r_2,r_1}$. In the extreme case that $r_1$ ignores $r_2$, while $r_2$ imitates $r_1$ perfectly, we will have $a_{r_1,r_2} = 1$ and $a_{r_2,r_1} = 0$. Note that both coefficients can be positive, in the case that $\eta_{r_1}$ and $\eta_{r_2}$ evolve smoothly and synchronously; indeed, such mutual connections appear frequently in the induced networks.

Equation~\ref{e:observation} specifies the \emph{observation} model, and Equation~\ref{e:dynamics} specifies the \emph{dynamics} model; together, they specify the joint probability distribution,
\begin{equation}
P(\eta, c  \mid  s; A, \sigma^2, \mu, \nu) = P(c \mid  \eta, s; \mu, \nu) P(\eta ; A),
\end{equation}
\change{where we omit subscripts to indicate the probability of all $\eta_{w,r,t}$ and $c_{w,r,t}$, given all $s_{r,t}, \mu_{r,t}, \nu_{w,t}$ and $A$.}

Because the observation model is non-Gaussian, the standard Kalman smoother cannot be applied. Inference under non-Gaussian distributions is often handled via second-order Taylor approximation, as in the extended Kalman filter~\cite{gelb1974applied}, but a second-order approximation to the Binomial distribution is unreliable when the counts are small. In contrast, sequential Monte Carlo sampling permits arbitrary parametric distributions for both the observations and system dynamics~\cite{Cappe2007Overview}.  Forward-filtering backward sampling~\cite{Godsill2004Monte} gives smoothed samples from the distribution $P(\eta_{w,1:R,1:T} \mid  c_{w,1:R,1:T}, s_{1:R,1:T}, A)$, so for each word $w$, we obtain a set of sample trajectories $\eta_{w,1:R,1:T}^{(k)}$, \change{where $k \in \{1,\ldots,K=100\}$ indexes the sample.} Monte Carlo approximation becomes increasingly accurate as $K \to \infty$~\cite{Cappe2007Overview}, but we found little change in the overall results for values of $K > 100$.

\subsubsection*{Inference and estimation} 
The total dimension of $\eta$ is equal to the product of the number of MSAs (200), words (2,603), and time steps (165), requiring inference over 85 million interrelated random variables.  To facilitate inference and estimation, we adopt a stagewise procedure. First we make estimates of the parameters $\nu$ (overall activation for each word) and $\mu$ (region-specific verbosity), assuming $\eta_{w,r,t} = 0, \forall w,r,t$. Next, we perform inference over $\eta$, assuming a simplified dynamics matrix $\tilde{A}$, which is diagonal. Last, we perform inference over the full dynamics matrix $A$, under $P(\eta)$; \change{this procedure is described in the next section}. See Figure~\ref{fig:block-diagram} for a block diagram of the inference and estimation procedure.

The parameters $\nu$ (global word activation) and $\mu$ (region-specific verbosity) are estimated first.
We begin by computing a simplified $\overline{\nu}_{w}$ as the inverse logistic function of the total frequency of word $w$, across all time steps. Next, we compute the maximum likelihood estimates of each $\mu_{r,t}$ via gradient descent. We then hold $\mu$ fixed, and compute the maximum likelihood estimates of each $\nu_{w,t}$. Inference over the latent spatiotemporal activations $\eta_{w,r,t}$ is performed via Monte Carlo Expectation Maximization (MCEM)~\cite{wei1990monte}.  For each word $w$, we construct a diagonal dynamics matrix $\tilde{A}_w$.  Given estimates of $\tilde{A}_w$ and $\sigma^2_w$, we use the sequential Monte Carlo
(SMC) algorithm of forward-filtering backward sampling
(FFBS)~\cite{Godsill2004Monte} to draw samples of $\eta_{w,1:R,1:T}$;
this constitutes the E-step of the MCEM process. Next, we apply
maximum-likelihood estimation to update $\tilde{A}_w$ and
$\sigma^2_w$; this constitutes the M-step.  These updates are
repeated until either the parameters converge or we reach a limit of
twenty iterations. We now describe each step in more detail:
\begin{itemize}
\item \textbf{E-step}.
The E-step consists of drawing samples from the posterior distribution over $\eta$. FFBS appends a backward pass to any SMC filter that produces a set of hypotheses and weights, \change{indexed by $k$. The weight $\omega^{(k)}_{w,r,t}$ represents the likelihood of the hypothesis $\eta^{(k)}_{w,r,t}$, so that the expected value $\mathbb{E}[\eta_{w,r,t}] = \frac{1}{\sum_k \omega^{(k)}_{w,r,t}}\sum_{k} \omega^{(k)}_{w,r,t} \eta^{(k)}_{w,r,t}$.} The role of the backward pass is to reduce variance by resampling the hypotheses according to the joint smoothing distribution.  Our forward pass is a standard bootstrap filter~\cite{Cappe2007Overview}: by setting the proposal distribution 
$q(\eta_{w,r,t} \mid \eta_{w,r,t-1})$ 
equal to the transition distribution 
$P(\eta_{w,r,t} \mid \mathbf{\eta}_{w,t-1}; A_w, \sigma^2_{w,r})$, 
the forward weights are equal to the recursive product of the observation likelihoods,
\begin{equation}
\omega^{(k)}_{w,r,t} = \omega^{(k)}_{w,r,t-1} P(c_{w,r,t} \mid  \eta_{w,r,t}, s_{w,t}; \nu_{w,t}, \mu_{r,t}).
\end{equation}

The backward pass uses these weights, and returns a set of unweighted hypotheses that are drawn directly from $P(\eta_{w,r,t} \mid  c_{w,r,t}, s_{r,t}; \nu_{w,t}, \mu_{r,t})$.
More complex SMC algorithms --- such as resampling, annealing, and
more accurate proposal distributions --- did not achieve higher
likelihood than the bootstrap filter.
\item \textbf{M-step}. The M-step consists of computing the average of
  the maximum likelihood estimates of $\tilde{A}_w$ and
  $\sigma^2_w$. Within each sample, maximum likelihood estimation is
  straightforward: the dynamics matrix $\tilde{A}_w$ is obtained by
  least squares, and $\sigma^2_{w,r}$ is set to the empirical variance
  $\frac{1}{T} \sum^T_t (\eta_{w,r,t} - \tilde{a}_{w,r}
  \eta_{w,r,t-1})^2.$
\end{itemize}

\subsubsection*{Examples}
Figure~\ref{f:example-etas} shows the result of this modeling procedure for several example words. In the right panel, each sample of $\eta$ is shown with a light dotted line.  In the left panel, the empirical word frequencies are shown with circles, and the smoothed frequencies for each sample are shown with dotted lines. Large cities generally have a lower variance over samples, because the variance of the maximum \emph{a posteriori} estimate of the binomial decreases with the total event count. For example, in Figure~\ref{f:example-etas}(c), the samples of $\eta$ are tightly clustered for Philadelphia (the sixth-largest MSA in the United States), but are diffuse
for Youngstown (the 95th largest MSA).  Note also that the relationship between frequency and $\eta$ is not monotonic --- for example, the frequency of \textit{ion} increases in Memphis over the duration of the sample, but the value of $\eta$ decreases. This is because of the parameter for background word activation, $\nu_{w,t}$, which increases as the word attains more general popularity. The latent variable model is thus able to isolate MSA-specific activation from nuisance effects that include the overall word activation and Twitter's changing sampling rate.

\subsection*{Constructing a network of linguistic diffusion}
Having obtained samples from the distribution $P(\eta \mid  c, s)$ over latent spatiotemporal activations, we now estimate the system dynamics, which describes the pathways of linguistic diffusion. Given the simple Gaussian form of the dynamics model (Equation~\ref{e:dynamics}), the coefficients $A$ can be obtained by ordinary least squares. We perform this estimation separately within each of the $K$ sequential Monte Carlo samples $\eta^{(k)}$, obtaining $K$ dense matrices $A^{(k)}$, for $k \in \{1, \ldots, K\}$. 

The coefficients of $A^{(k)}$ are not in meaningful units, and their relationship to demographics and geography will therefore be difficult to interpret, model, and validate. Instead, we prefer to use a binarized, \emph{network} representation, $\mathbb{B}$. Given such a network, we can directly compare the properties of linked MSAs with the properties of randomly selected pairs of MSAs not in $\mathbb{B}$, offering face validation of the proposed link between macro-scale linguistic influence and the demographic and geographic features of cities.

Specifically, we are interested in a set of pairs of MSAs, $\mathbb{B} = \{ \langle r_1,r_2 \rangle \}$, for which we are confident that $a_{r_1,r_2} > 0$, given the uncertainty inherent in estimation across sparse word counts. Monte Carlo inference enables this uncertainty to be easily quantified: we compute $z$-scores $z_{r_1,r_2}$ for each ordered city pair, using the empirical mean and standard deviation of $a^{(k)}_{r_1,r_2}$ across samples $k \in \{1, \ldots, K\}$. We select pairs whose $z$-score exceeds a threshold $z^{(\text{thresh})}$, denoting
the selected set
$\overline{\mathbb{B}} = \{\langle r_i, r_j \rangle : z_{i,j} > z^{(\text{thresh})}\}$.
To compute uncertainty around a large number of coefficients,
we apply the Benjamini-Hochberg False Discovery Rate (FDR) correction for multiple hypothesis testing~\cite{benjamini1995controlling}, which controls the expected proportion of false positives in $\overline{\mathbb{B}}$ as
\begin{equation}
 \text{FDR}(z^{(\text{thresh})}) 
 = \frac
    {P_{null}(z_{i,j} > z^{(\text{thresh})})}
    {\tilde{P}(z_{i,j} > z^{(\text{thresh})})}
 = \frac
    {1-\Phi(z^{(\text{thresh})})}
    { [R(R-1)]^{-1} \sum_{i \neq j} 1\{z_{i,j} > z^{(\text{thresh})}\} },
\end{equation}
\change{where $P_{null}$ is the probability, under a one-sided hypothesis, that $z$ exceeds $z^{(\text{thresh})}$}
under a standard normal distribution, which we would expect if $a_{i,j}$ values were random;
this has probability $1-\Phi(z^{(\text{thresh})})$, where $\Phi$ is the Gaussian CDF.
$\tilde{P}$ is the simulation-generated empirical distribution 
over $z(a_{i,j})$ values.
If high $z$-scores occur much more often under the model ($\tilde{P}$)
than we would expect by chance ($P_{null}$), only a small proportion should be expected to be false positives; the Benjamini-Hochberg ratio
is an upper bound
on the expected proportion of false positives in $\mathbb{B}$.
To obtain $\text{FDR} < 0.05$, the individual test threshold is approximately $z^{(\text{thresh})} = 3.2$, or in terms of $p$-values, $p < 6 \times 10^{-4}$. 
We see 510 dynamics coefficients survive this threshold; these indicate high-probability pathways of linguistic diffusion. 
The associated set of city pairs is denoted $\overline{\mathbb{B}}_{0.05}$.

Figure~\ref{f:map} shows a sparser network $\overline{\mathbb{B}}_{0.001}$, induced using a more stringent threshold of $\text{FDR} < 0.001$. The role of geography is apparent from the figure: there are dense connections within regions such as the Northeast, Midwest, and West Coast, and relatively few cross-country connections.  For example, we observe many connections among the West Coast cities of San Diego, Los Angeles, San Jose, San Francisco, Portland, and Seattle (from bottom to top on the left side of the map), but few connections from these cities to other parts of the country.

\paragraph{Practical details} To avoid overfitting and degeneracy in the estimation of $A^{(k)}$, we place a zero-mean Gaussian prior on each element $a^{(k)}_{r_1,r_2}$, tuning the variance $\lambda$ by grid search on the log-likelihood of a held-out subset of time slices within $\eta_{1:T}$. The maximum \emph{a posteriori} estimate of $A$ can be computed in closed form via ridge regression. Lags of length greater than one are accounted for by regressing the values of $\eta_t$ against the moving average from the previous ten time steps. Results without this smoothing are
broadly similar.

\subsection*{Geographic and demographic correlates of linguistic diffusion}
By analyzing the properties of pairs of metropolitan areas that are connected in the network $\mathbb{B}$, we can quantify the geographic and demographic drivers of online language change. Specifically, we construct a logistic regression to identify the factors that are associated with whether a pair of cities have a strong linguistic connection. The positive examples are pairs of MSAs with strong transmission coefficients $a_{r_1,r_2}$; an equal number of negative examples is sampled randomly from a distribution $Q$, which is designed to maintain the same empirical distribution of MSAs as appears in the positive examples. This ensures that each MSA appears with roughly the same frequency in the positive and negative pairs, eliminating a potential confound. 

The independent variables in this logistic regression include  geographic and demographic properties of pairs of MSAs. We include the following demographic attributes: median age, log median income, and the proportions of, respectively, African Americans, Hispanics, individuals who live in urbanized areas, and individuals who rent their homes. The proportion of European Americans was omitted because of a strong negative correlation with the proportion of African Americans; the proportion of Asian Americans was omitted because it is very low for the overwhelming majority of the 200 largest MSAs. These raw attributes are then converted into both asymmetric and symmetric predictors, using the raw difference and its absolute value. The symmetric predictors indicate pairs of cities that are likely to share influence; besides the demographic attributes, we include the geographical distance. The asymmetric predictors are properties that may make an MSA likely to be the driver of online language change. Besides the raw differences of the six demographic attributes, we include the log difference in population. All variables are standardized. 

For a given demographic attribute, a negative regression coefficient for the absolute difference would indicate that similarity is important; a positive regression coefficient for the (asymmetric) raw difference would indicate that regions with large values of this attribute tend to be senders rather than receivers of linguistic innovations.  \change{For example, a strong negative coefficient for the asymmetric log difference in population would indicate that larger cities usually lead smaller ones, as proposed in the gravity and cascade models.}

To visually verify the geographic distance properties of our model,
Figure~\ref{f:shuf_vs_real} compares networks obtained by discretizing $A^{(k)}$ against networks of randomly-selected MSA pairs, sampled from $Q$. Histograms of these distances are shown in Figure~\ref{f:shuf_vs_real_histograms}, and their average values are shown in Table~\ref{t:features}. The networks induced by our model have many more short-distance connections \change{than would be expected by chance}. Table~\ref{t:features} also shows that many other demographic attributes are more similar among cities that are linked in our model's network. 

A logistic regression can show the extent to which each of the above predictors relates to the dependent variable, the binarized linguistic influence. However, the posterior uncertainty of the estimates of the logistic regression coefficients depends not only on the number of instances (MSA pairs), but principally on the variance in the Monte Carlo-based estimates for $A^{(k)}$, which in turn depends on the sampling variance and the size of the observed spatiotemporal word counts. To properly account for this complex variance, we run the logistic regression separately within each Monte Carlo sample $k$, and report the empirical standard errors of the logistic coefficients across the samples.

\paragraph{Practical details} This procedure requires us to discretize the dynamics network \emph{within} each sample, which we will write $\mathbb{B}^{(k)}$. One solution would be simply take the $L$ largest values; alternatively, we could take the $L$ coefficients for which we are most confident that $a^{(k)}_{r_1,r_2} > 0$. We strike a balance between these two extremes by sorting the dynamics coefficients according to the lower bound of their 95\% confidence intervals. This ensures that we get city pairs for which $a^{(k)}_{r_1,r_2}$ is significantly distinct from zero, but that we also emphasize large values rather than small values with low variance. Per-sample confidence intervals are obtained by computing the closed form solution to the posterior distribution over each dynamics coefficient, $P(a_{r_1,r_2}^{(k)} \mid  \eta_{r_1}^{(k)}, \eta_{r_2}^{(k)}, \lambda)$, which, in ridge regression, is normally distributed. We can then compute the 95\% confidence interval of the coefficients in each $A^{(k)}$, and sort them by the bottom of this confidence interval, $\tilde{a}^{(k)}_{i,j} = \mu_{a^{(k)}_{i,j}} - Z_{(.975)} \sigma^2_{a^{(k)}_{i,j}}$, where $Z_{(.975)}$ is the inverse Normal cumulative density function evaluated at 0.975, $Z_{(.975)} = 1.96$.
We select $L$ by the number of coefficients that pass the $p < 0.05$ false discovery rate threshold in the aggregated network ($L=510$), as described in the previous section. This procedure yields $K = 100$ different discretized influence networks $\mathbb{B}^{(k)}$, each with identical density to the aggregated network $\overline{\mathbb{B}}$. By comparing the logistic regression coefficients obtained within each of these $K$ networks, it is possible to quantify the effect of uncertainty about $\eta$ on the substantive inferences that we would like to draw about the diffusion of language change. 

\section*{Results} 
Figure~\ref{f:log-reg-symm} shows the resulting logistic regression coefficients. While geographical distance is prominent, the absolute difference in the proportion of African Americans is the strongest
predictor: the more similar two metropolitan areas are in terms of this demographic, the more likely that linguistic influence is transmitted between them. Absolute difference in the proportion of Hispanics, residents of urbanized areas, and median income are also strong predictors. This indicates that while language change does spread geographically, demographics play a central role, and nearby cities may remain linguistically distinct if they differ demographically, particularly in terms of race.  \change{In spoken language}, African American English differs more substantially from other American varieties than any regional dialect~\cite{wolfram2005american}; our analysis suggests that such differences persist in the virtual and disembodied realm of social media. Examples of linguistically linked city pairs that are \change{geographically} distant but demographically similar include Washington D.C. and New Orleans (high proportions of African-Americans), Los Angeles and Miami (high proportions of Hispanics), and Boston and Seattle (relatively few minorities, compared with other large cities).

Of the asymmetric features, population is the most informative, as
larger cities are more likely to transmit to smaller ones. 
In the induced network of linguistic influence $\overline{\mathbb{B}}_{0.05}$, the three largest
metropolitan areas -- New York, Los Angeles, and Chicago -- have 40
outgoing connections and only fifteen incoming connections. \change{These findings are in accord with theoretical models offered by Trudgill~\cite{trudgill1974linguistic} and Labov~\cite{labov2003pursuing}.} Wealthier and younger cities are also significantly more likely to lead than to follow. While this may seem to conflict with earlier findings that language change often originates from the working class, wealthy \emph{cities} must be differentiated from wealthy \emph{individuals}: wealthy cities may indeed be the home to the upwardly-mobile working class that Labov associates with linguistic creativity~\cite{labov2001principles}, even if they also host a greater-than-average number of very wealthy individuals.

Additional validation for the logistic regression is obtained by
measuring its cross-validated predictive accuracy. For each of the $K$
samples, we randomly select 10\% of the instances (positive or
negative city pairs) as a held-out test set, and fit the logistic
regression on the other 90\%.  For each city pair in the test set, the
logistic regression predicts whether a link exists, and we check the
prediction against whether the directed pair is present in $\mathbb{B}^{(k)}$.  Results are shown in Table~\ref{tab:predictive}. Since the number of positive and negative instances are equal, a random baseline would achieve 50\% accuracy. A classifier that uses only geography and population (the two components of the gravity model) gives 66.5\% predictive accuracy. The addition of demographic features (both asymmetric and symmetric) increases this substantially, to 74.4\%. While symmetric features obtain the most robust regression coefficients, adding the asymmetric features increases the predictive accuracy from 74.1\% to 74.4\%, a small but statistically significant difference.

\section*{Discussion}
Language continues to evolve in social media. By tracking the
popularity of words over time and space, we can harness large-scale data to uncover the hidden structure of language change. We find a remarkably strong role for demographics, particularly as our analysis is centered on a geographical grouping of individual users. Language change is significantly more likely to be transmitted between demographically-similar areas, especially with regard to race --- although demographic properties such as socioeconomic class may be more difficult to assess from census statistics.

Language change spreads across social network connections, and it is well known that the social networks that matter for language change are often strongly homophilous in terms of both demographics and geography~\cite{milroy1991language,labov2001principles}. This paper approaches homophily from a macro-level perspective: rather than homophily between individual speakers~\cite{kwak2010twitter}, we identify homophily between geographical communities as an important factor driving the observable diffusion of lexical change. Individuals who are geographically proximate will indeed be more likely to share social network connections~\cite{sadilek2012finding}, so the role of geography in our analysis is not difficult to explain. But more surprising is the role of demographics, since it is unclear whether individuals who live in cities that are geographically distant but demographically similar will be likely to share a social network connection. Previous work has shown that friendship links on Facebook are racially homophilous~\cite{chang2010epluribus}, but to our knowledge the interaction of urban demographics with geography has not been explored. In principle, a large-scale analysis of social network links on Twitter or some other platform could shed light on this question. Such \change{platforms} impose restrictions that make social networks difficult to acquire, but one possible approach would be to try to link the ``reply trees'' considered by Gon\c{c}alves et al~\cite{gonccalves2011modeling} with the geographic and demographic metadata considered here; while intriguing, this is outside the scope of the present paper. A methodological contribution of our paper is the demonstration that similar macro-scale social phenomena can be inferred directly from spatiotemporal word counts, even without access to individual social networks.

Our approach can be refined in several ways. We gain robustness by choosing metropolitan areas as the basic units of analysis, but measuring word frequencies among sub-communities or individuals could shed light on linguistic diversity \emph{within} metropolitan areas. Similarly, estimation is facilitated by fitting a single first-order dynamics matrix across all words, but some regions may exert more or less influence for different types of words, and a more flexible model of temporal dynamics might yield additional insights. Finally, language change occurs at many different levels, ranging from orthography to syntax and pragmatics. This work pertains only to word frequencies, but future work might consider structural changes, such as the phonetical process resulting in the transcription of \textit{i don't} into \textit{ion}.

It is inevitable that the norms of written language must change to accommodate the new ways in which writing is used. As with all language changes, innovation must be transmitted between real language users, ultimately grounding out in countless individual decisions --- conscious or not --- about whether to use a new linguistic form. 
Traditional sociolinguistics has produced many insights from the close analysis of a relatively small number of variables. Analysis of large-scale social media data offers a new, complementary methodology by aggregating the linguistic decisions of millions of individuals.

\section*{Acknowledgments}
We thank
Jeffrey Arnold, 
Chris Dyer, 
Lauren Hall-Lew, 
Scott Kiesling, 
Iain Murray, 
John Nerbonne, 
Bryan Routledge, 
Lauren Squires, 
and Anders S{\o}gaard 
for comments on this work.


\clearpage

\section*{Supporting Information Legends}

\begin{description}
\item[Appendix S1. Term list.] List of all words considered in our main analysis.
\item[Appendix S2. Term examples.] Examples for each term considered in our analysis.
\item[Appendix S3. Data Procedures.] Description of the procedures used for data processing, including Twitter data acquisition, geocoding, content filtering, word filtering, and text processing.
\item[Table S1. Term annotations.] Tab-separated file describing annotations of each term as entities, foreign-language, or acceptable for analysis.
\item[Software S1. Preprocessing software.] Source code for data preprocessing.
\end{description}

\clearpage
\section*{Figures}
 \input{figures_content}

\clearpage
\section*{Tables}

\begin{table}[!ht]
\centering   \begin{tabular}{@{\vrule height 10.5pt depth4pt  width0pt}ccc}
  & mean & st. dev \\
  \hline
  Population & 1,170,000 & 2,020,000 \\
  Log Population & 13.4 & 0.9 \\
  \% Urbanized & 77.1 & 12.9\\
  Median Income & 61,800 & 11,400\\
  Log Median Income & 11.0 & 0.2\\
  Median age & 36.8 & 3.9\\
  \% Renter     & 34.3 & 5.2\\
  \% Af. Am      & 12.9 & 10.6\\
  \% Hispanic & 15.0 & 17.2\\
  \hline
\end{tabular}
\caption{
\textbf{Statistics of metropolitan statistical areas.} 
Mean and standard deviation for demographic attributes of the 200 Metropolitan Statistical Areas (MSAs) considered in our study. 
}
\label{tab:avg-demo}
\end{table}

\begin{table}[!ht]
  \centering
  \begin{tabular}{lp{5.5in}}
    $c_{w,r,t}$ & Number of individuals who used word $w$ in metropolitan area $r$ during week $t$. \\
    $s_{r,t}$ & Number of individuals who posted messages in metropolitan area $r$ at time $t$.\\
    $p_{w,r,t}$ & Empirical probability that an individual from metropolitan area $r$ will use word $w$ during week $t$.\\
    $\eta_{w,r,t}$ & Latent spatiotemporal activation for word $w$ in metropolitan area $r$ at time $t$. \\
    $\nu_{w,t}$ & Global activation for word $w$ at time $t$.\\
    $\mu_{r,t}$ & Regional activation (``verbosity'') for metropolitan  area $r$ at time $t$.\\
    $a_{r_1,r_2}$ & Autoregressive coefficient from metropolis $r_1$ to $r_2$.\\
    $A = \{a_{r_1,r_2}\}$ & Complete autoregressive dynamics matrix. \\
    $\sigma^2_{w,r}$ & Autoregressive variance for $\eta_{w,r,t}$, for all times $t$.\\
    $\lambda$ & Variance of zero-mean Gaussian prior over each $a_{r_1,r_2}$.\\
    $\omega^{(k)}_{w,r,t}$ & Weight of sequential Monte Carlo hypothesis $k$ for word $w$, metropolis $r$, and time $t$.\\
    $z_{r_1,r_2}$ & $z$-score of $a_{r_1,r_2}$, computed from empirical distribution over Monte Carlo samples.\\
    $\overline{\mathbb{B}}$ & Set of ordered city pairs for whom $a_{r_1,r_2}$ is significantly greater than zero, computed over all samples.\\
    $\mathbb{B}^{(k)}$ & Top $L$ ordered city pairs, as sorted by the bottom of the 95\% confidence interval on $\{ a^{(k)}_{r_1,r_2} \}$.\\
    $Q$ & Random distribution over discrete networks, designed so that the marginal frequencies for ``sender'' and ``receiver'' metropolises are identical to their empirical frequencies in the model-inferred network.
  \end{tabular}
  \caption{\textbf{Table of mathematical notation.}}
  \label{tab:notation}
\end{table}

\begin{table}[!ht]
\centering
\begin{tabular}{@{\vrule height 10.5pt depth4pt  width0pt}lllll}
 & linked mean & linked s.e. & nonlinked mean & nonlinked s.e.\\ 
 \hline
\emph{geography}\\
distance (km) &  $919$ &  $36.5$ &  $1940$ &  $28.6$\\ 
\emph{symmetric}\\
abs diff \% urbanized &  $9.09$ &  $0.246$ &  $13.2$ &  $0.215$\\ 
abs diff log median income &  $0.163$ &  $0.00421$ &  $0.224$ &  $0.00356$\\ 
abs diff median age &  $2.79$ &  $0.0790$ &  $3.54$ &  $0.0763$\\ 
abs diff \% renter     &  $4.72$ &  $0.132$ &  $5.38$ &  $0.103$\\ 
abs diff \% af. am      &  $6.19$ &  $0.175$ &  $14.7$ &  $0.232$\\ 
abs diff \% hispanic &  $10.1$ &  $0.375$ &  $20.2$ &  $0.530$\\ 
\emph{asymmetric}\\
raw diff log population &  $0.247$ &  $0.0246$ &  $-0.0127$ &  $0.00961$\\ 
raw diff \% urbanized &  $1.77$ &  $0.389$ &  $-0.0912$ &  $0.112$\\ 
raw diff log median income &  $0.0320$ &  $0.00654$ &  $-0.00166$ &  $0.00187$\\ 
raw diff median age &  $-0.198$ &  $0.113$ &  $-0.00449$ &  $0.0296$\\ 
raw diff \% renter     &  $0.316$ &  $0.195$ &  $-0.00239$ &  $0.0473$\\ 
raw diff \% af. am      &  $0.00292$ &  $0.244$ &  $0.00712$ &  $0.109$\\ 
raw diff \% hispanic &  $0.0327$ &  $0.472$ &  $0.0274$ &  $0.182$\\ 
\hline
\end{tabular}
\caption{Differences between linked and (sampled) non-linked pairs of cities, summarized by their mean and its standard error.} \label{t:features} 
\end{table}

\begin{table}[!ht]
\centering
\begin{tabular}{lcc}
 & mean acc & std.~err\\ 
 \hline 
geography + symmetric + asymmetric &  $74.37$ &  $0.08$\\ 
geography + symmetric &  $74.09$ &  $0.07$\\ 
symmetric + asymmetric &  $73.13$ &  $0.08$\\ 
geography + population &  $67.33$ &  $0.08$\\ 
geography &  $66.48$ &  $0.09$\\ 
\hline 
\end{tabular}
\caption{Average accuracy predicting links between MSA pairs, and its Monte Carlo standard error (calculated from $K=100$ simulation samples). The feature groups are defined in Table~\ref{t:features}; ``population'' refers to ``raw diff log population.''
\label{tab:predictive}}
\end{table}
\end{document}

%% file: figures_content.tex

\begin{figure}[!ht]
  \begin{center}
 \includegraphics[clip=true,trim=1in 0.8in 0 .5in,width=5.5in]{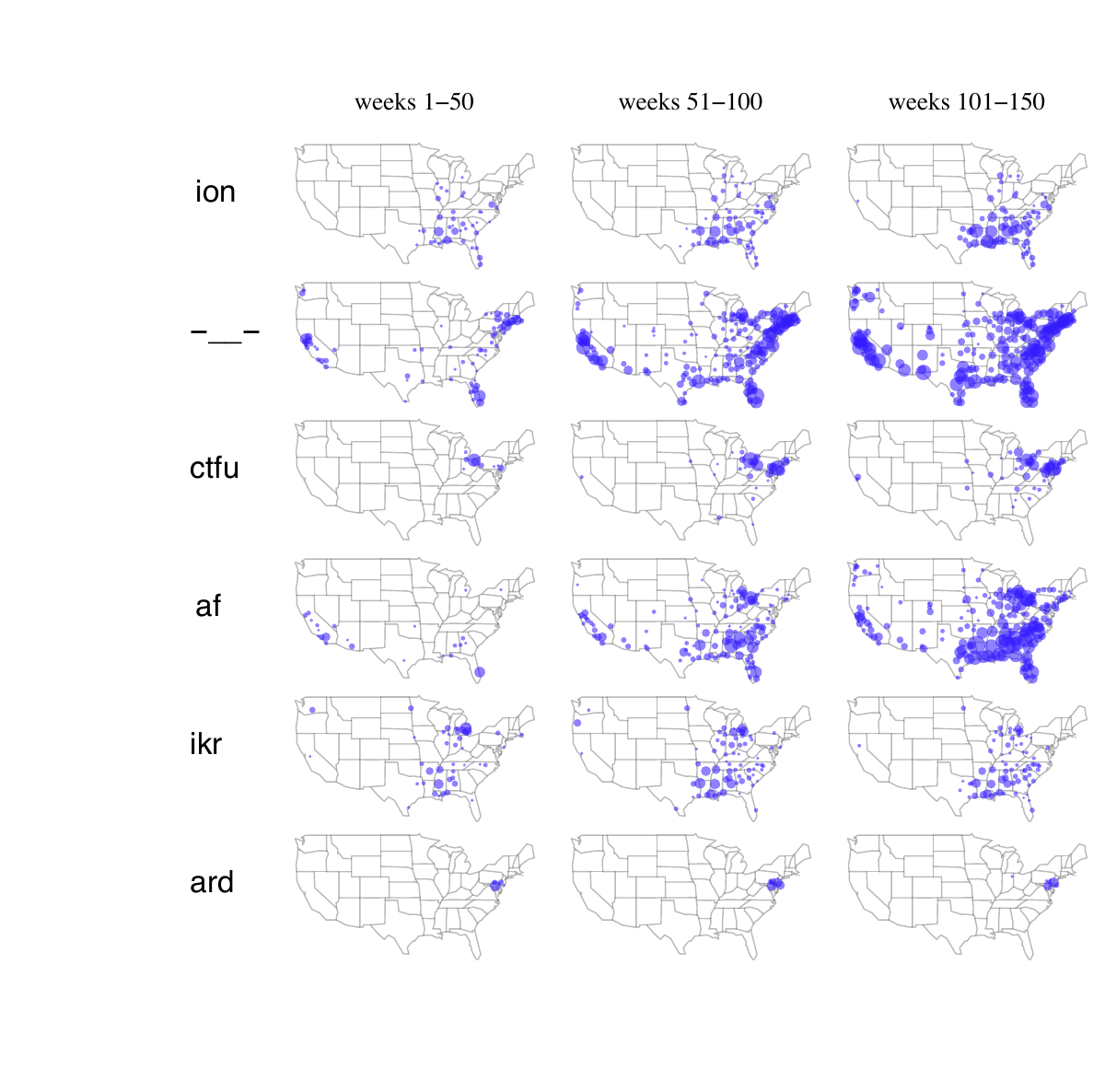}  
\end{center}
  \caption{\textbf{Change in frequency for six words: 
    \textit{ion}, \textit{-\_\_-}, \textit{ctfu},
    \textit{af}, \textit{ikr}, \textit{ard}}.  
  Blue circles indicate cities where on average, at least 0.1\% of users use the word during a week. A circle's area is proportional to the word's probability.
\label{f:ex-bruh}
  } 
\end{figure}

\begin{figure}[!ht]
  \centering
  \includegraphics[width=5.5in]{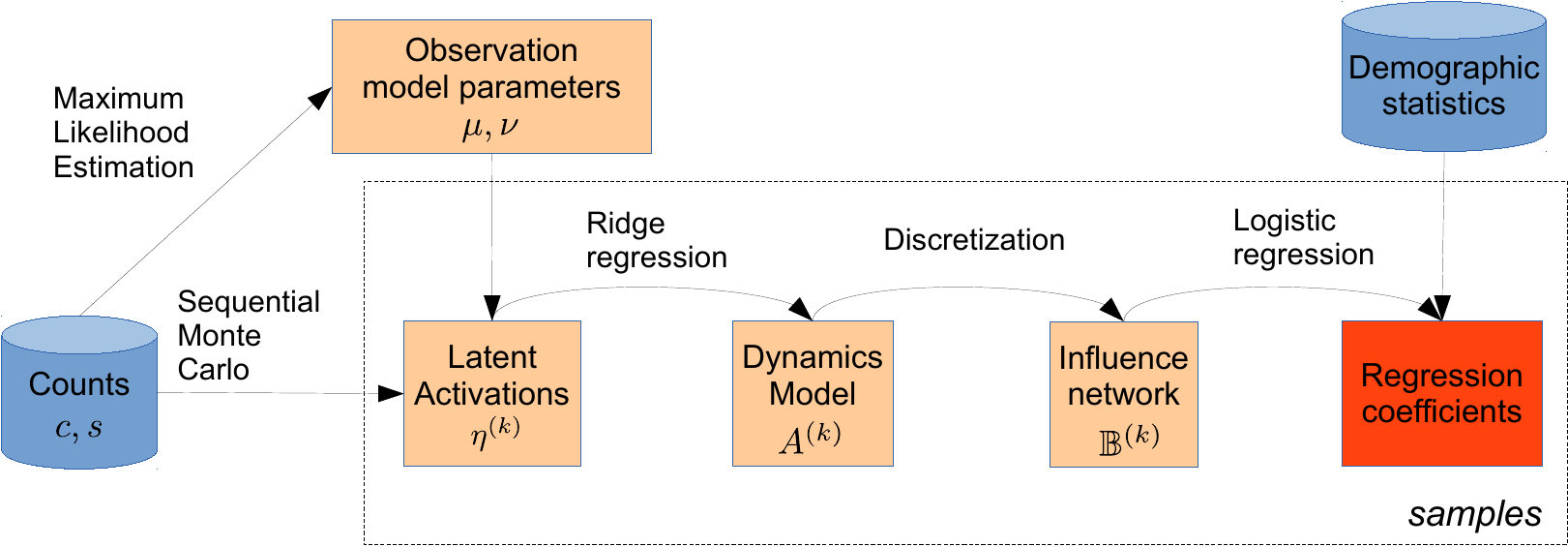}
  \caption{\textbf{Block diagram for our statistical modeling procedure}. The dotted outline indicates repetition across samples drawn from sequential Monte Carlo.
  \label{fig:block-diagram}
  } 
\end{figure}

\begin{figure}[!ht]
  \vspace{-.2in}
  \begin{tabular}{lcm{3.9cm}}
  (a) & \qquad \textit{ion} \qquad & \includegraphics[width=4in]{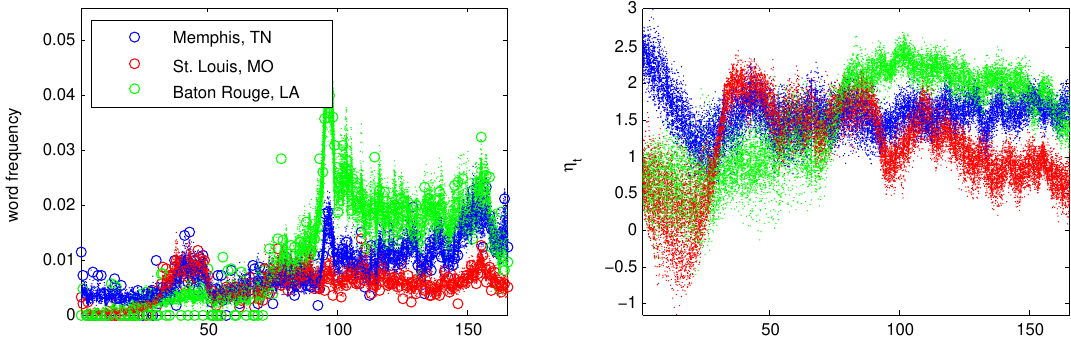} \\
  (b) & \qquad \textit{-\_\_-} \qquad & \includegraphics[width=4in]{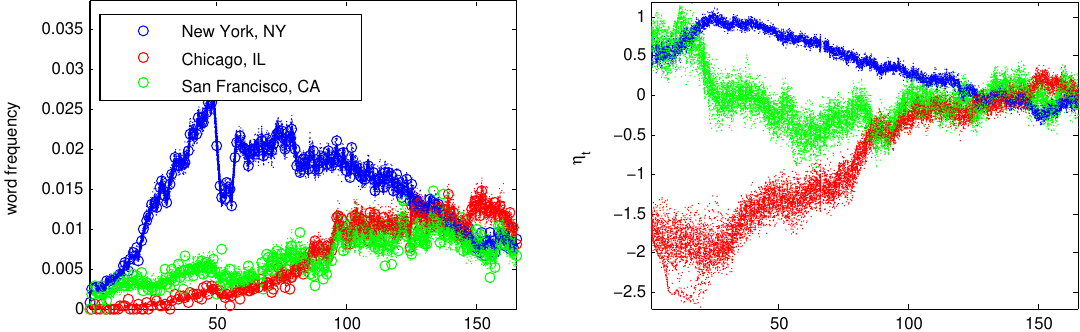} \\
  (c) & \qquad \textit{ctfu} \qquad & \includegraphics[width=4in]{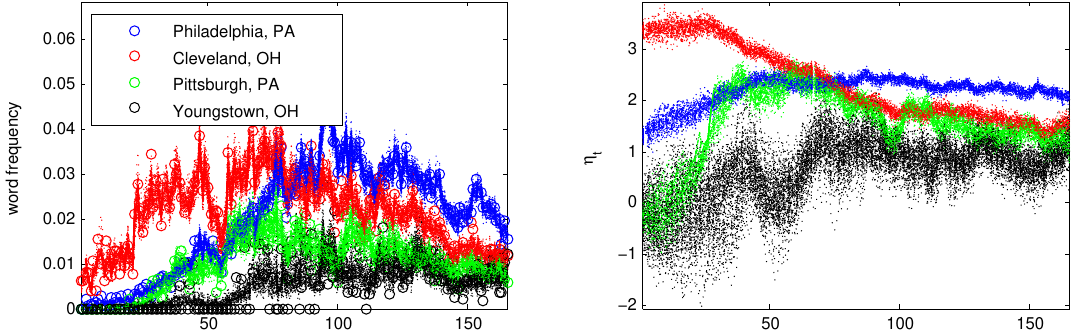}\\
  (d) & \qquad \textit{af} \qquad & \includegraphics[width=4in]{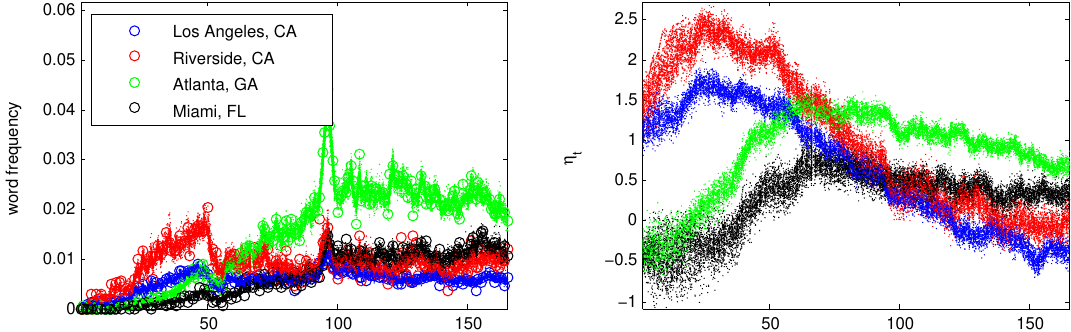} \\
  (e) & \qquad \textit{ikr} \qquad & \includegraphics[width=4in]{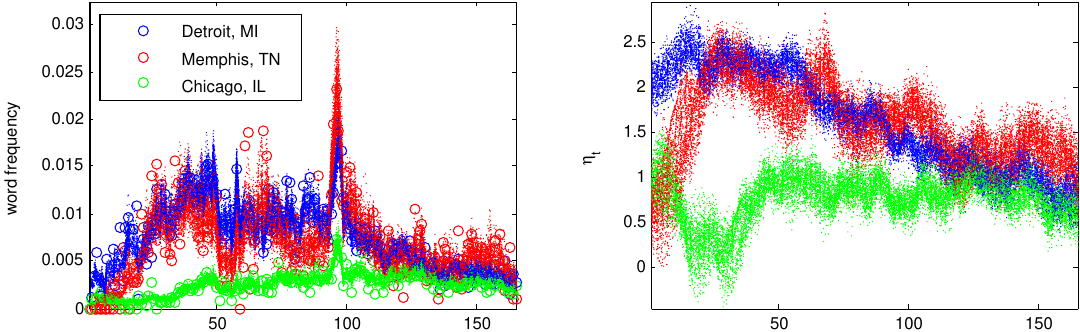} \\
  (f) & \qquad \textit{ard} \qquad & \includegraphics[width=4in]{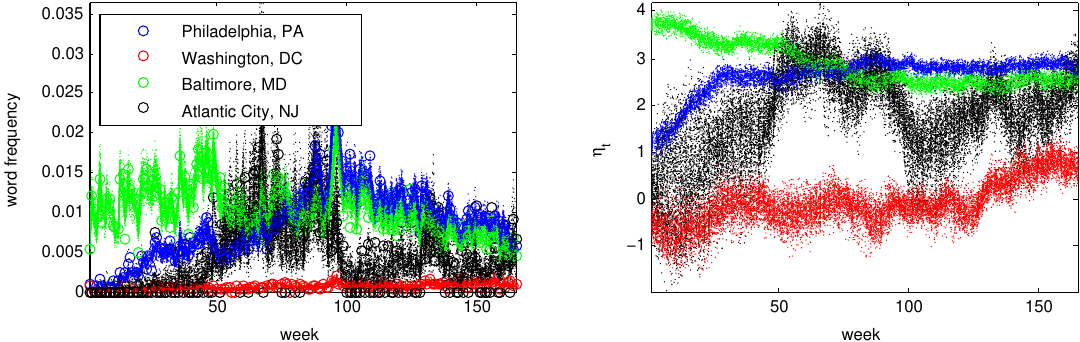} \\
\end{tabular}
\caption{\textbf{Left: empirical term frequencies (circles) and their Monte Carlo smoothed estimates (dotted lines); Right: Monte Carlo smoothed estimates of $\eta$.}
\label{f:example-etas}
} 
\end{figure}

\begin{figure}[!ht]
  \begin{center}
  \includegraphics[width=4.5in]{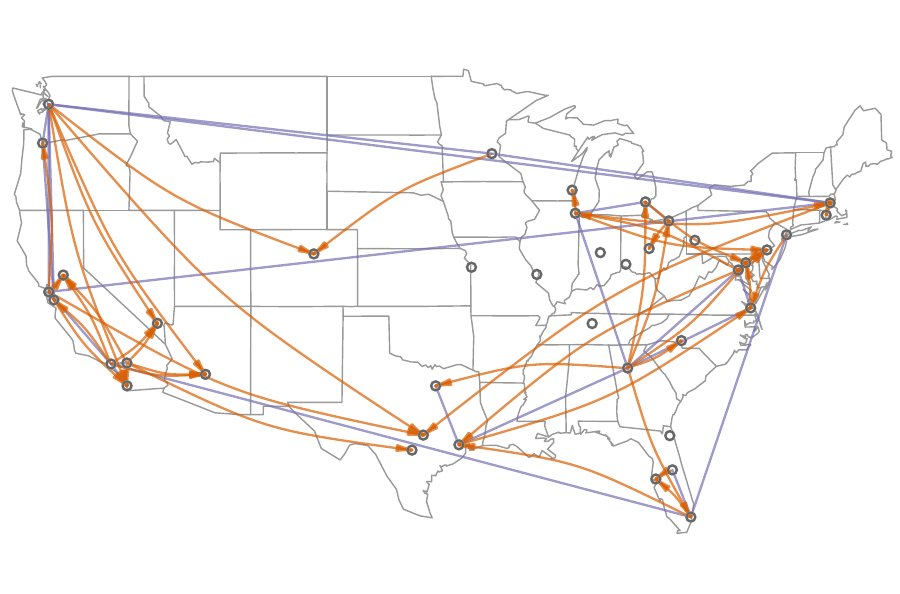}
  \caption{\textbf{Induced network, showing significant
    coefficients among the 40 most populous MSAs (using an
    $\text{FDR}<0.001$ threshold, yielding 254 links)}.  
    Blue edges represent bidirectional
    influence, when there are directed edges in both
    directions; orange links are unidirectional.
    \label{f:map}
  } 
\end{center}
\end{figure}

\begin{figure}[!ht]
\begin{center}
\includegraphics[width=3in]{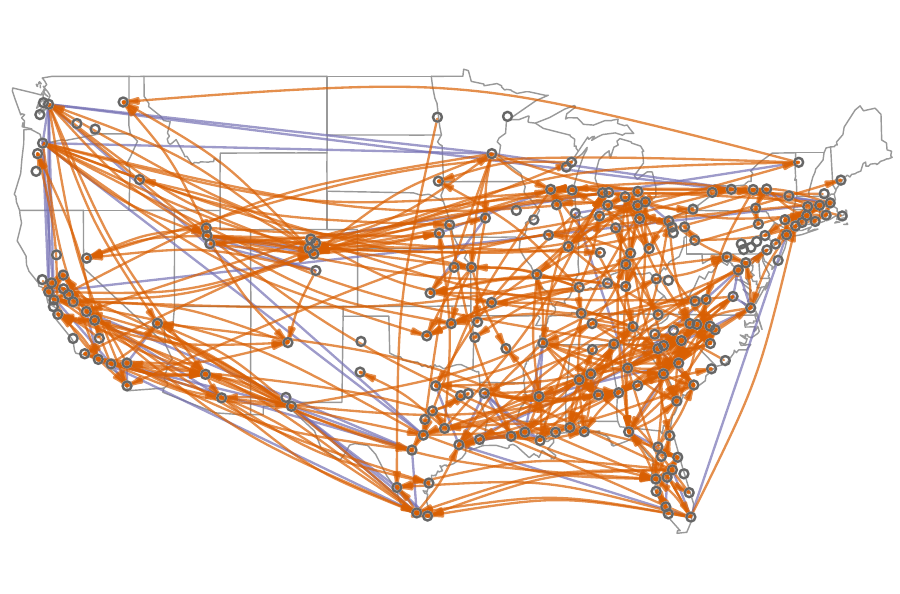}
\includegraphics[width=3in]{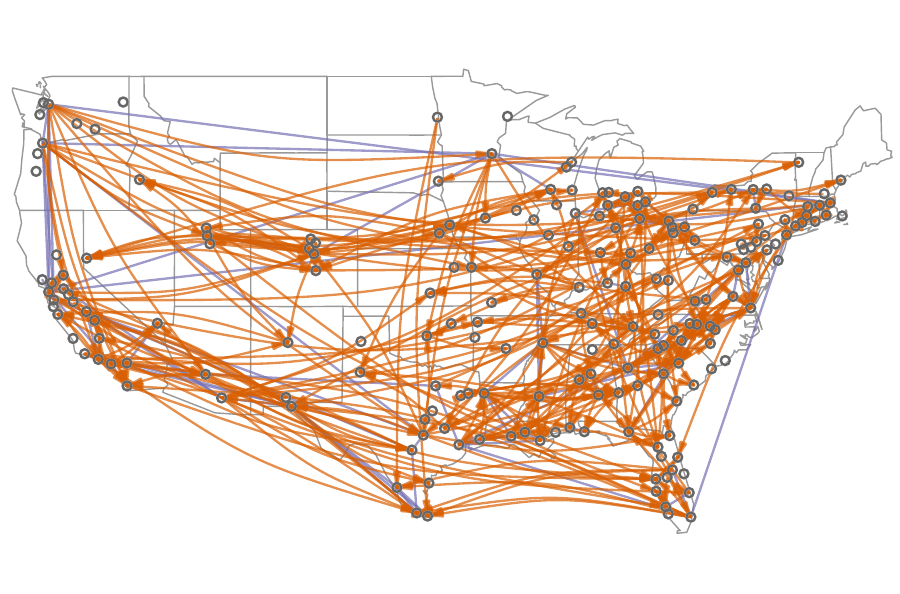}
\noindent\rule{6in}{0.4pt}
\includegraphics[width=3in]{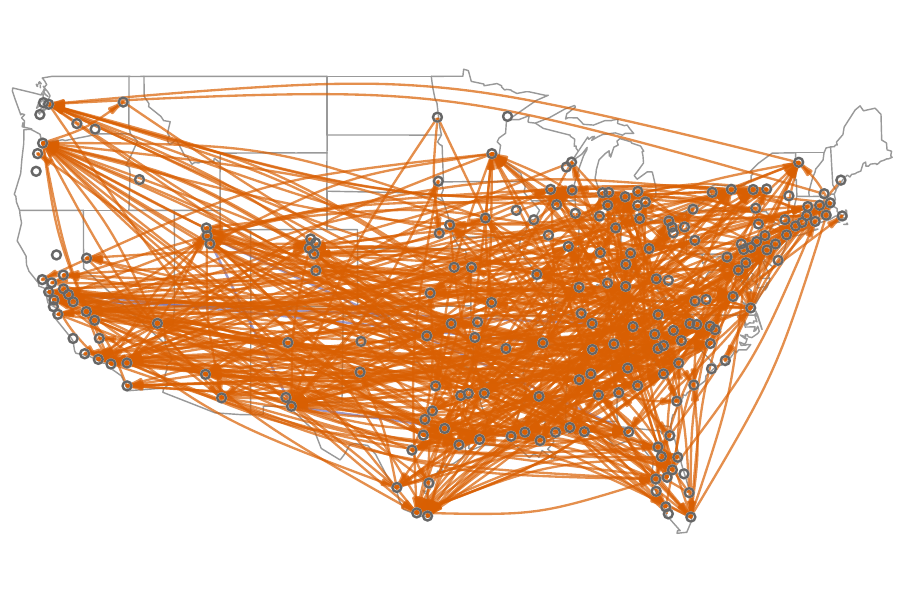}
\includegraphics[width=3in]{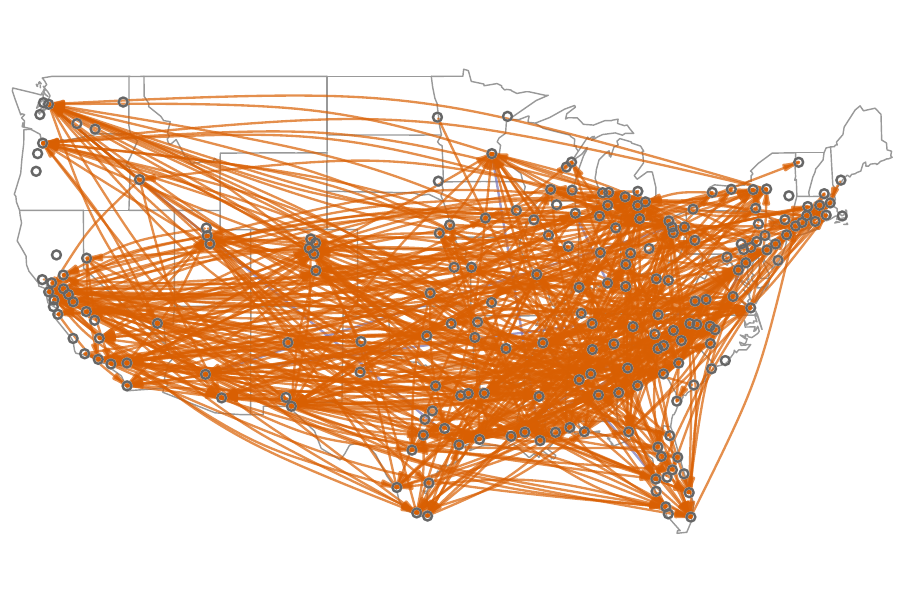}
\caption{\textbf{Top: two sample networks inferred by the model, $\mathbb{B}_{0.05}$}. (Unlike Figure~\ref{f:map}, all 200 cities are shown.) 
\textbf{Bottom: two ``negative'' networks, sampled from $Q$}; these are samples from the non-linked pair distribution $Q$, which is constructed to have the same marginal distributions over senders and receivers as in the inferred network. A blue line indicates directed edges in both directions between the pair of cities; orange lines are unidirectional.
\label{f:shuf_vs_real}
} 
\end{center}
\end{figure}

\begin{figure}[ht!]
\begin{center}
\includegraphics[width=3in]{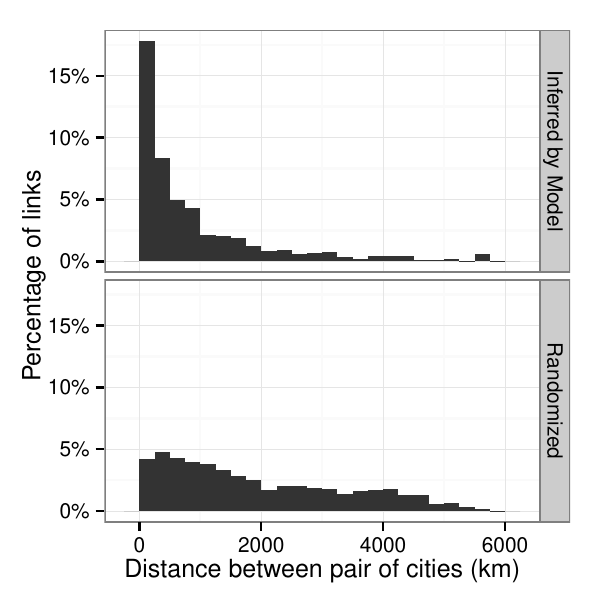}
\caption{
\textbf{Histograms of distances between pairs of connected cities}, in model-inferred networks (top), versus ``negative'' networks from $Q$ (bottom).
\label{f:shuf_vs_real_histograms}
} 
\end{center}
\end{figure}

\newpage

\begin{figure}[ht!]
\centering
\includegraphics[clip=true,trim=0in 0in 1in 0,width=6.5in]{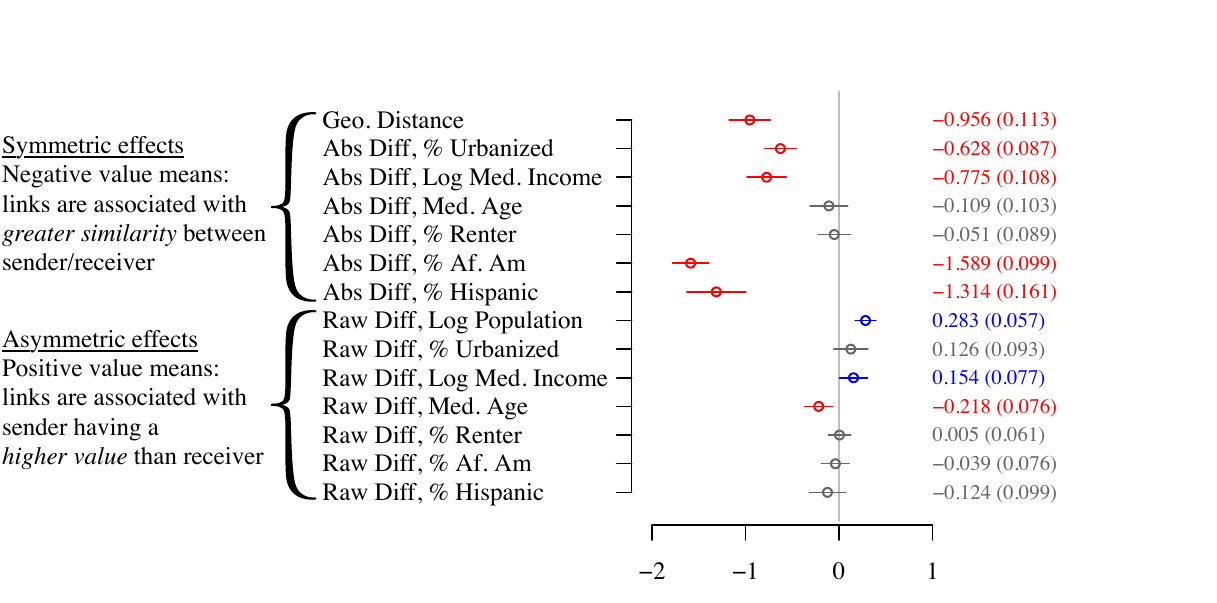}
\caption{\textbf{Logistic regression coefficients for predicting links between city (MSA) pairs}.  95\% confidence intervals are plotted; standard errors are in parentheses.  Coefficient values are from standardized inputs; the mean and standard deviations are shown to the right.
  \label{f:log-reg-symm}
} 
\end{figure}